\documentclass[letterpaper, 10 pt, journal, twoside]{IEEEtran}

\usepackage[numbers]{natbib}
\usepackage{multicol}
\usepackage[bookmarks=true]{hyperref}

\usepackage{graphicx}
\usepackage[caption=false]{subfig}
\usepackage[nolist,nohyperlinks]{acronym}
\renewcommand{\baselinestretch}{0.92}

\begin{document}

\title{Exploring Probabilistic Distance Fields in Robotics}

\author{Lan Wu. Email: Lan.Wu-2@uts.edu.au \\
Robotics Institute, University of Technology Sydney }



%

\markboth{Robotics: Science and Systems (RSS) conference, Pioneer Workshop, 2024}
{Robotics: Science and Systems (RSS) conference, Pioneer Workshop, 2024} 

\maketitle



\section{Introduction}\label{sec:introduction}
A Robot's successful performance of a highly intelligent mission typically entails the integration of several foundational research tasks. 
Each task imposes unique requirements on the task-oriented representation, reflecting robotic representations' diversity and complexity. 
For example, feature-based representations~\cite{orb_features,surf,sift} are widely used by most SLAM methods for localisation and mapping tasks~\cite{paper:orbslam2,vins-mono} and by computer vision for the context of Structure from Motion~\cite{ozyecsil2017SurveyStructureFromMotion,MVE}. However, it is difficult to reconstruct a dense map from sparse features for visualisation tasks and to use sparse landmarks for planning and human interaction. Raw and dense representations provide decent visual information and high texture in geometry via a large point cloud~\cite{rusu2008towards,RTAB-MAP}. The drawback is they usually require managing and storing a large amount of data. Moreover, they give a low-level understanding of the scene, which is of limited applicability for navigation tasks. Volumetric Dense Representations~\cite{voxel1989,voxel1996,tsdf,kinectfusion,TSDFplanning1,TSDFplanning2} such as \ac{TSDF} are more suitable for planning, obstacle avoidance, manipulation, and reconstruction. However, without matching sparse features to identify correspondences, it is difficult to perform a consistent alignment for localisation~\cite{reijgwart2020voxgraph}. 

Thus, designing a representation for every different task of a complex mission is impractical and costly. Unified and task-centric representations that fulfil multiple requirements remain unexplored, demanding more innovation to shape the future of robotic intelligence. The most relevant representations that aim to cater for multiple tasks are as follows. Compared to \ac{TSDF}, \ac{ESDF}~\cite{Voxblox,pan2022voxfield,zhu2021vdbEDT,han2019fiesta} contains distance even with gradient fields over the entire exploring environment suitable for localisation, mapping, planning and manipulation. A major challenge is that it is expensive to compute and propagate the distance information over the whole space when a task might only require this information at discrete sparse points.
Recent developments with neural networks include a continuous implicit representation for reconstruction~\cite{gropp2020learning_shapes,park2019deepsdf,mildenhall2021nerf,sitzmann2020implicit,mildenhall2021nerf,mescheder2019occupancy_networks}. 
Such neural representations have shown great promise for mapping, localisation~\cite{sucar2021imap,rosinol2022nerf_slam,zhu2022nice_slam} as well as planning~\cite{ortiz2022isdf,driess2022learningSDF_planning,adamkiewicz2022planning_using_NeRFs,pantic2022NeRF_planning}. Although these neural representations serve as a sound representation for multiple tasks, they are a `black-box' which requires further theoretical formulation. 

On the other hand, Gaussian Process (GP)~\cite{GPbook,gp-gpis-opt,paper:GeomPrior,paper:GPImplSurf, GPOccMaps} probabilistic representation is a well-established approach that deals with the sparseness of sensory information, manages and handles uncertainty, allows a continuous prediction and even supports high-level semantic understanding of the scene. I believe GP-based probabilistic representations have the potential to offer a theoretical sound, universal and "one for all" representation solution for localisation, visualisation, mapping, motion planning, manipulation, human-robot interaction and many more demands in robotics. 

\begin{figure}[t]
	\centering
	\resizebox{1\linewidth}{!}{
	\includegraphics[]{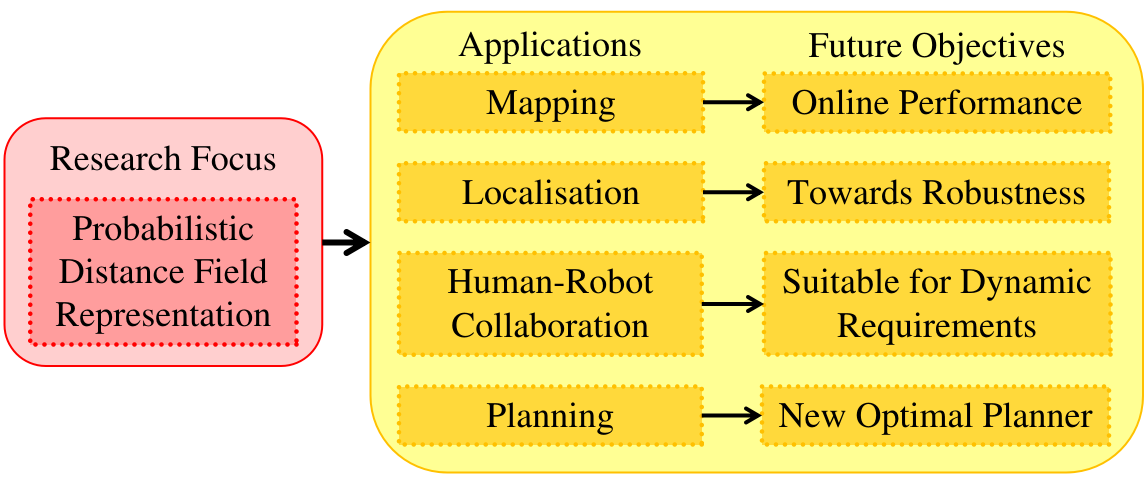}
	}
	\caption{Block Diagram of the proposed GP-based representation architecture for multiple robotics applications and future research objectives.}
\vspace{-2ex}
	\label{fig:diagram}
\end{figure}

As shown in Fig.~\ref{fig:diagram}, my work proposes a GP-based probabilistic distance field (GPDF) representation that mathematically models the fundamental property of Euclidean distance field (EDF) along with gradients, surface normals and dense reconstruction~\cite{wu2023probabilistic}. EDF with gradients is crucial for motion planning and dynamic obstacle avoidance in human-robot interaction tasks. Surface normal directly feeds tasks such as grasping. A densely reconstructed mesh can be used for visualisation and graphical interfaces. The accurately inferred distance fields allow pose estimation via direct optimisation of the alignment. The distance and gradient are only computed when tasks are querying the representation. Furthermore, the nature of the GP representation provides uncertainty for probabilistic data fusion. Overall, GPDF overcomes the limitations of other state-of-the-art representations in robotics, as we highlighted above. GPDF offers a unified solution that transcends the need to design multiple representations for diverse underlying tasks. 

I anticipate that GPDF will form the basis representation of the full robot autonomy.
GPDF can be extended for efficient online running, more robust performance, for tightly integrating with a new motion planner taking advantage of all the information output from GPDF, and for applying it to the human-robot collaboration and manipulation tasks. GPDF focus on spatial-only information at the current stage. Future vision considers expanding it to include temporal information or even equipped with scene understanding. The nature of the GP has the advantage of integrating GPDF with Gaussian spatting for high-quality reconstruction. Currently, GPDF works with depth sensors. The probabilistic nature of GPDF can be directly used for data fusion, combining multiple sensors in the future for spatial, temporal and semantic information.
In what follows, I will present the progress to support the prospection throughout my PhD and first-year postdoc.

\section{Progress to Date}


\subsection{Probabilistic Distance Fields}
I initially proposed GPDF as a novel representation~\cite{wu2021faithful}. My key contribution is the realisation that the regularised Eikonal equation can be simply solved by applying the logarithmic transformation to a GP formulation to approximate the accurate Euclidean distance fields. The proposed approach does not require post-processing or wave propagation to recover the distance fields. Our experiments show that it produces the most accurate results for distance and surface reconstruction. However, it requires a trade-off between accuracy and interpolation abilities.
These are limiting factors for specific applications that require highly accurate values of the Euclidean distance fields, such as in the case of echolocation for robotic inspection. 
Additionally, due to the non-linear transformation that is applied over the GP fields, the variance prediction of our proposed method makes it hard to model the field's uncertainty reliably.
Following the theoretical formulation in~\cite{wu2021faithful}, we have introduced the so-called reverting GP distance field~\cite{le2023accurate} that increased the accuracy and provided a proxy for its uncertainty. GP-based distance fields are appealing as they can represent complex environments with non-parametric models. The accuracy of the proposed distance fields enables us to tackle the multiple key problems in robotics with noisy measurements and uncertainty.

\subsection{Unified Representation}
My work proposed GPDF as a unified representation for the multiple robotic tasks in~\cite{wu2023mop}. It allows us to use a single representation for localising, mapping, and path planning. The continuous and probabilistic nature provides an accurate dense surface for mapping. 
The accurately inferred distance fields allow pose estimation via direct optimisation of the alignment between the global map representation and a new set of measurements. 
Furthermore, GPDF is integrated with a optimisation-based planners. 

\subsection{Towards Efficiency of the Probabilistic Representation}
I have been working on efficient GP modelling. Firstly, I proposed a novel submapping method~\cite{LanRAL20}, following the topology of the scene to generate conditional independent submaps of GP. Large datasets can efficiently be processed by the proposed pipeline, producing not only a surface but also the uncertainty information of the representation. Secondly, the pseudo inputs applied in efficient GP frameworks are a subset of the full data in general. We proposed an alternative solution~\cite{wu2023pseudo} to constrain the locations of pseudo data using the distance field itself provided by our GP kernel functions. We demonstrate the effectiveness and sound performance of our method in different applications.

\section{Ongoing and Future Work}
Developing and maintaining multiple representations tailored to each task requires significant time, computational resources, and expertise. In contrast, adopting a unified GPDF representation approach offers a more streamlined and cost-effective solution.
I believe that GPDF serves as the cornerstone for robots to accomplish more complex and challenging tasks. 
By leveraging GPDF, robots can navigate through intricate environments, understand spatial relationships, and interact with objects and humans seamlessly. 
Toward this objective, some ongoing and future works are highlighted. 

\subsection{Online Operation of Probabilistic Distance Fields}\label{sec:online operation}
The main concern with GP-based algorithms is the cubic computational complexity~\cite{GPbook}. Substantial efforts have been made in the literature and my progress to date. However, a notable gap remains to achieve the desired online performance. I am developing an online GPDF framework that provides distance, gradients, repulsive information, rapid continuous reconstruction and colour as output. Our framework develops a fast access data structure to maintain feasible, long-term and large-sized mapping and exploration. It is important to have online GPDF  since most robotic tasks have to be conducted in real-time. Our ongoing progress (in Fig.~\ref{fig:onlineGP}) has successfully generated varying information in real-time and is more efficient than similar state-of-the-art frameworks~\cite{Voxblox,pan2022voxfield}.

\subsection{Capability for Human-Robot Collaboration}\label{sec:HRI}
\begin{figure}[ht]
  \centering
  \resizebox{0.40\linewidth}{!}{
  \subfloat[\label{fig:onlineGP}]
  {\includegraphics[height=2cm]{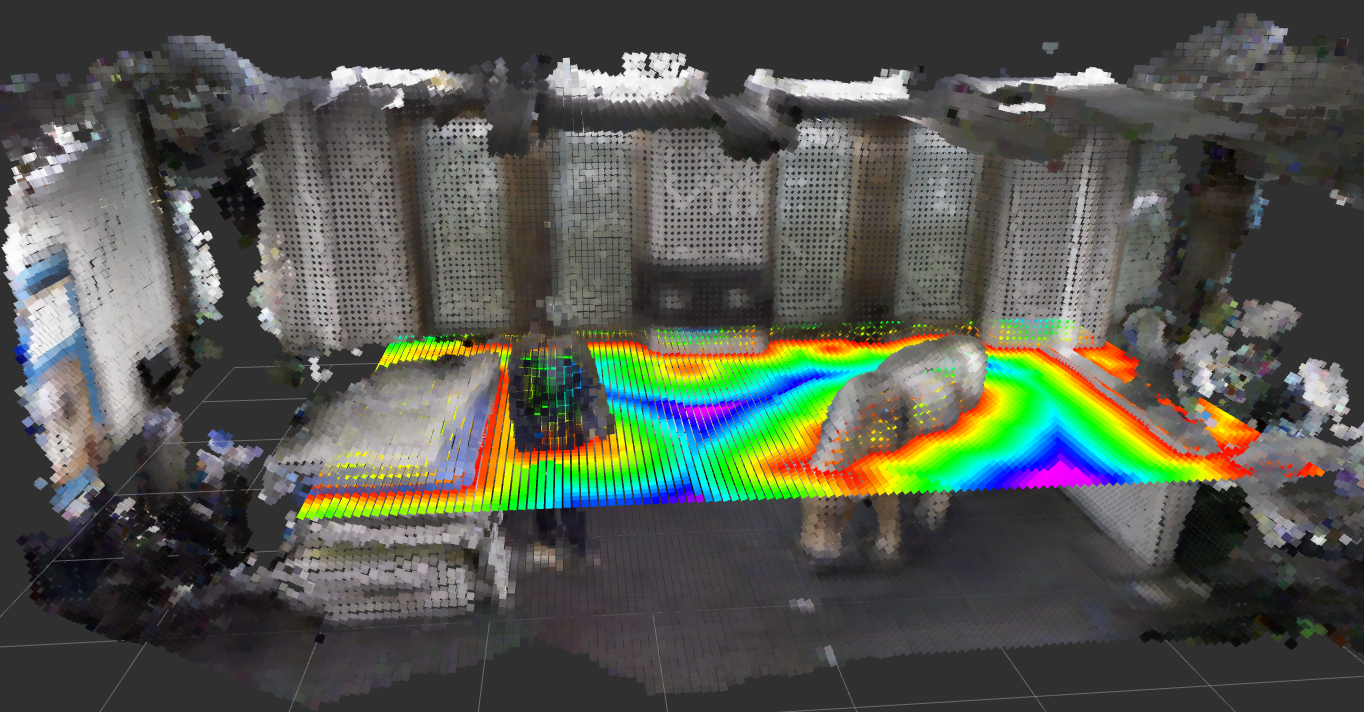}}}
  \resizebox{0.58\linewidth}{!}{
  \subfloat[\label{fig:HRI}]
  {\includegraphics[height=3cm]{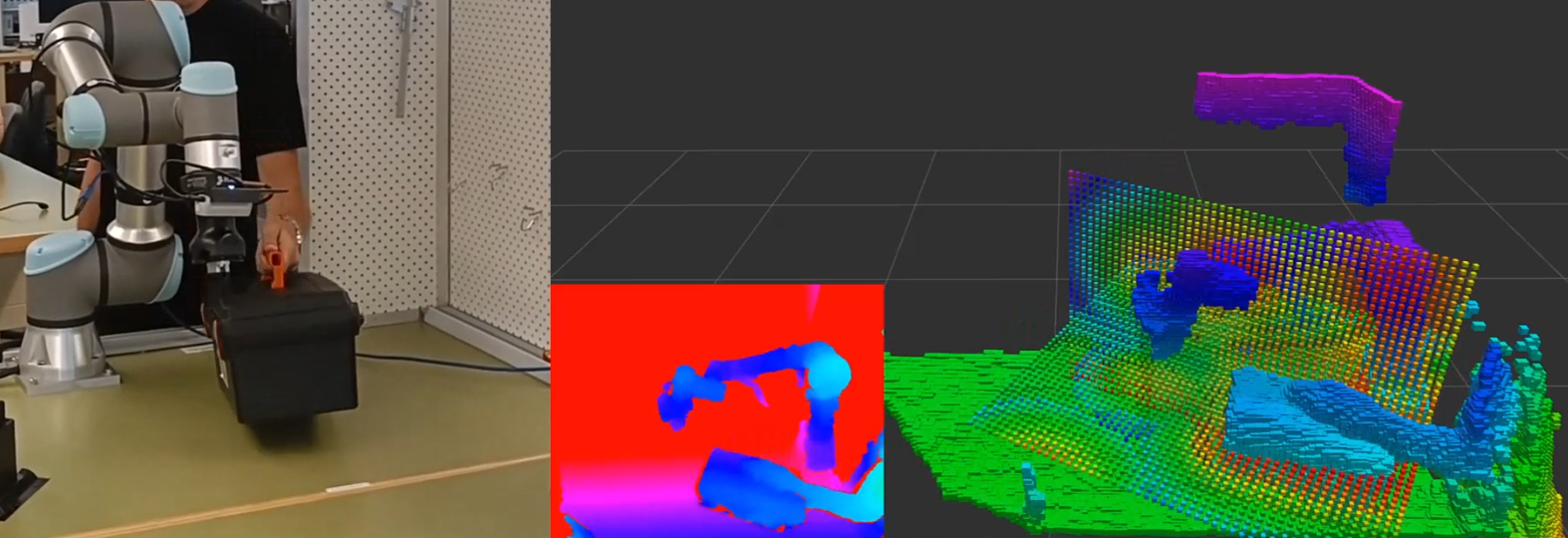}}}
  \caption{a) Online GPDF framework performance for Sec.~\ref{sec:online operation}. b) Dynamical GPDF for Human-robot collaboration application for Sec.~\ref{sec:HRI}.}
  \vspace{-2ex}
\end{figure}

We are extending the GPDF for Human-Robot Collaboration (HRC) and grasping tasks. One essential requirement for the perception of the interactive scene is to deal with dynamic objects and human interaction in the area. My previous research outcomes have yielded progress in dynamic application~\cite{liu2021active,wu2023pseudo}. However, further work is needed to bridge the remaining distance to attain the interactive distance fields with effective dynamic capability and efficiency. As shown in Fig.~\ref{fig:HRI}, we have open-sourced our framework~\cite{ali2024interactive} for HRC.\footnote[3]{\tt \url{https://uts-ri.github.io/IDMP}}.

\subsection{Robust Performance for Localisation}
I am developing the machine perception part for a project delivering a form of technologically enhanced human echolocation that enables blind users to perceive their surroundings in precise detail. One requirement is to solve the well-known SLAM problem~\cite{slam_problem,Cadena16tro-SLAMfuture}. One of my future works is to combine feature-based odometry with our GPDF-based odometry~\cite{wu2023mop} to create a more robust localisation framework. The solution needs to be computationally light and robust, enabling it to run smoothly in an embedded device with AI glasses. 
%
\subsection{Theoretical New Planner Based on GP Distance Fields}
With various information directly queried from the GPDF, other applications will take advantage of the novel representation solution. We are working on designing a new planning algorithm with distance, gradient and repulsive information. 

\subsection{Vision for Future Work}
Currently, the primary focus of GPDF lies in spatial representation. However, our future vision entails expanding its capabilities to include temporal information and scene understanding. The nature of GP has the potential to integrate with Gaussian Splatting for radiance field rendering and be used for probabilistic data fusion in future applications. 


\begin{acronym}[AAAAAA]

    \acro{UTS}{University of Technology Sydney }
    \acro{RI}{Robotics Institute }
    \acro{FEIT}{Faculty of Engineering and Information Technology}
    \acro{CAS}{Centre for Autonomous Systems }
    \acro{WiEIT}{Women in Engineering and IT}
    \acro{1D}{One-Dimensional }
    \acro{2D}{Two-Dimensional }
    \acro{2.5D}{Two-and-a-Half-Dimensional}
    \acro{3D}{Three-Dimensional }
    \acro{GP}{Gaussian Process }
    \acro{GPIS}{Gaussian Process Implicit Surfaces }
    \acro{MAP}{Maximum A Posteriori }
    \acro{MLE}{Maximum Likelihood Estimation }
    \acro{ICP}{Iterative Closest Point }
    \acro{MVE}{Multi-View Environment }
    \acro{OG}{Occupancy Grid }
    \acro{CHOMP}{Covariant Hamiltonian Optimization for Motion Planning}
    \acro{FIESTA}{Fast Incremental
Euclidean DiSTAnce Fields}
    \acro{GPUs}{Graphics Processing Units }
    \acro{KD}{K-Dimensional}
    
    \acro{Log-GPIS}{Log-Gaussian Process Implicit Surfaces }
    \acro{LiDAR}{Light Detection And Ranging Sensor }
    \acro{SLAM}{Simultaneous Localisation and Mapping }
    \acro{TDF}{Truncated Distance Field }
    \acro{EDF}{Euclidean Distance Field }
    \acro{SDF}{Signed Distance Field }
    \acro{TSDF}{Truncated Signed Distance Field }
    \acro{ESDF}{Euclidean Signed Distance Field }
    
    \acro{GPIS-SDF}{GPIS with signed distance function }
    
    \acro{RGB}{Red-Green-Blue }
    \acro{RGB-D}{Red-Green-Blue-Depth }
    \acro{RMSE}{Root Mean Squared Error }
    
    \acro{CI}{Conditional Independent }
    \acro{FP}{Forward Propagation}
    \acro{BP}{Backward Propagation}
    \acro{D-SKI}{Structured Kernel Interpolation framework with Derivatives}
    \acro{SKI}{Structured Kernel Interpolation method}
    \acro{D-SKI-CI-Fusion}{Structured Kernel Interpolation with Derivatives and Conditional Independent Fusion}
    \acro{MVMs}{Matrix-Vector Multiplications method}

    \acro{PDE}{Partial Differential Equation }
    \acro{Log-GPIS-MOP}{Log-Gaussian Process Implicit Surface for Mapping, Odometry and Planning }
    \acro{RANSAC}{Random Sample Consensus}
    \acro{Dynamic-GPDF}{Dynamic Gaussian Process Distance Fields}
\end{acronym}

\bibliographystyle{IEEEtran}
\bibliography{reference}

\end{document}